%% file: iclr2027_conference.tex
\documentclass{article}

\usepackage{iclr2027_conference,times}
\input{math_commands.tex}

\usepackage{hyperref}
\usepackage{url}
\usepackage{graphicx}
\usepackage{booktabs}
\usepackage{multirow}
\usepackage[table]{xcolor}

\definecolor{citationblue}{RGB}{35,80,150}
\definecolor{internalred}{RGB}{150,55,45}
\definecolor{urlblue}{RGB}{35,100,150}

\hypersetup{
    colorlinks=true,          
    citecolor=citationblue,   
    linkcolor=internalred,    
    urlcolor=urlblue          
}

\title{DiFuse: Fusing Native Diffusion Representations for Spatial Reasoning in Unified Multimodal Models}

\author{
Ruixin Fan$^{1}$ \quad
Weining Wang$^{1}$ \quad
Jiahui Sun$^{1}$ \quad
Tailai Chen$^{1}$ \quad
Jing Liu$^{1}$\\[1ex]
$^{1}$ Institute of Automation, Chinese Academy of Sciences
}

\iclrfinalcopy

\begin{document}
\maketitle

\fancyhead{}
\fancyhead[L]{Preprint}

\begin{abstract}
Unified multimodal models (UMMs) jointly support visual understanding and image generation, yet information flow between these two capabilities remains largely asymmetric: understanding representations routinely condition generation, whereas generative representations are rarely reused for multimodal reasoning.
In this work, we investigate whether intermediate representations from a UMM's native diffusion branch can enhance spatial reasoning without compromising general multimodal understanding.
We propose \textbf{DiFuse}, which selectively incorporates intermediate diffusion representations into the understanding pathway of UMMs. 
Through systematic layer-wise analysis, we identify which diffusion representations are most informative and where they can be integrated with minimal interference to semantic processing. Guided by these findings, DiFuse spatially aligns diffusion features with visual tokens and injects them into selected multimodal decoder layers via residual fusion.
Extensive experiments on UniWorld-V1 show that, DiFuse consistently improves spatial reasoning over a matched SFT baseline while maintaining broadly comparable performance on general multimodal benchmarks.
Further analyses show that these gains depend critically on the choice of diffusion representations, injection positions, and the preservation of spatial correspondence between diffusion features and visual tokens.
Results on OmniGen2 further demonstrate the transferability of DiFuse across unified multimodal architectures. 
Overall, our findings show that native diffusion representations can provide complementary spatial cues for multimodal reasoning when selectively integrated in a structure-preserving manner.

\end{abstract}

\begin{figure}[h]
    \centering
    \includegraphics[width=1.0\textwidth]{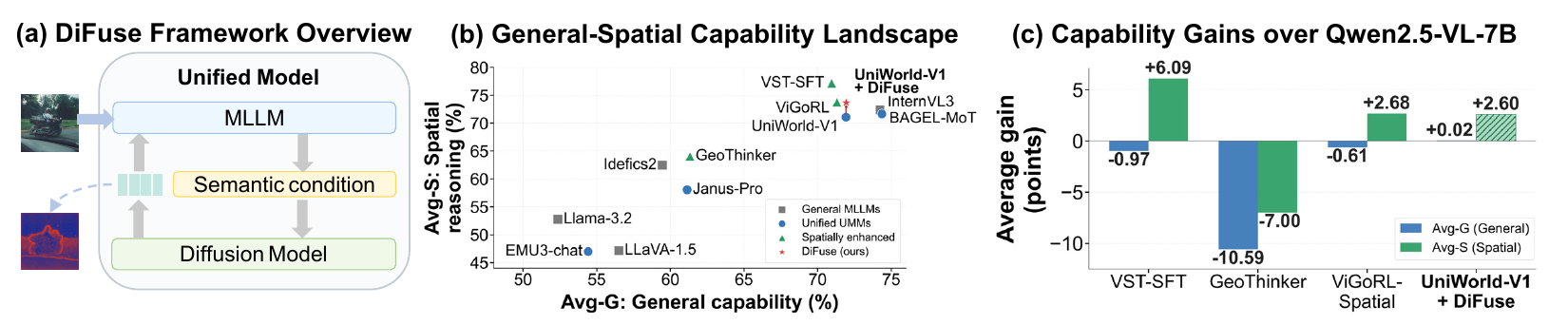}
    \caption{\textbf{Overview of DiFuse and its impact on unified multimodal models.} 
    (a) DiFuse establishes a generation-to-understanding pathway by reusing intermediate representations from the native Diffusion Transformer (DiT) branch and integrating them into the visual understanding pathway of a unified multimodal model. (b) General–spatial capability landscape. Models are positioned by their average general-understanding (Avg-G) and spatial-reasoning (Avg-S) scores. Marker shapes distinguish general-purpose MLLMs, UMMs, spatially enhanced models, and DiFuse. (c) Gains relative to Qwen2.5-VL-7B. We compare four Qwen2.5-VL-7B-based models against the same baseline. Blue and green bars show changes in Avg-G and Avg-S, respectively. UniWorld-V1 + DiFuse improves Avg-S by 2.60 points while leaving Avg-G nearly unchanged (+0.02 points).}
    \label{fig:overview}
\end{figure}

\section{Introduction}
\label{sec:intro}
Multimodal understanding and image generation have traditionally developed along largely separate trajectories. Recent unified multimodal models (UMMs) bring these capabilities into a shared architecture, enabling a single model to both interpret and generate visual content \citep{team2024chameleon, xie2025show, wang2024emu3, lin2025uniworld, deng2025emerging, wu2025omnigen2}. Despite this architectural unification, however, the interaction between understanding and generation remains markedly asymmetric: representations from the understanding branch are routinely used to condition generation, whereas representations learned by the generative branch are rarely reused for understanding.
Recent studies have begun to explore generation-to-understanding transfer through diffusion feedback, reconstruction objectives, or auxiliary generation tasks \citep{luo2025deem, wang2025diffusion, ma2025genhancer, su2026generation}.
However, these approaches primarily transfer generative knowledge through additional optimization objectives or intermediate generation processes.
Whether representations already learned by the native generative branch of a UMM can be directly reused to improve its understanding capability remains largely unexplored. In this work, we study this reverse information flow and ask: Can native diffusion representations improve spatial reasoning without compromising general multimodal understanding?

Spatial reasoning provides a natural setting for studying this question because the understanding and generation branches are optimized for different representational objectives.
Vision-language encoders are optimized for semantic alignment and invariance, which support robust high-level recognition but may reduce sensitivity to local spatial relationships and geometric structure \citep{radford2021learning, zhai2023sigmoid, tong2024eyes}. In contrast, diffusion transformers learn to reconstruct visual content under noise perturbations and therefore retain information about object layout, boundaries, and spatial organization \citep{rombach2022high, peebles2023scalable, fuest2026diffusion}. 
Recent geometry-enhanced multimodal models and diffusion-based spatial reasoning methods further suggest that such structural representations can complement semantic visual features \citep{zheng2026learning, huang20263drs, fan2026vlm, li2026thinking, wu2026generation}.

However, these approaches often introduce structural information through dedicated geometry modules or external generative models. Their effectiveness can further depend on feature selection, injection position, fusion strategy, and spatial correspondence. We therefore investigate how diffusion representations can be integrated into the native understanding pathway without disrupting the original semantic pathway. 
However, effectively transferring diffusion representations to understanding is non-trivial. Intermediate diffusion features vary across layers and denoising stages, and naive fusion may introduce irrelevant information or disrupt semantic representations. Moreover, preserving spatial correspondence between diffusion features and visual tokens is essential for transferring local structural cues. Therefore, the key challenge is to determine which diffusion representations are useful, where they should be integrated, and how their spatial structure should be preserved.

To address these questions, we conduct a systematic study within unified multimodal architectures. We first perform a layer-wise analysis to determine which intermediate diffusion representations are useful and where they can be integrated with minimal interference to the original semantic pathway.
Guided by this analysis, we propose DiFuse, a post-training framework that directly incorporates intermediate representations from the native Diffusion Transformer (DiT) into the understanding pathway.
Figure~\ref{fig:overview} illustrates the core idea: the diffusion model supplies intermediate DiT features, which DiFuse injects into the MLLM to complement the UMM's existing semantic pathway.
Experiments on UniWorld-V1 show that DiFuse consistently improves spatial reasoning over a matched SFT baseline while maintaining broadly comparable general multimodal performance. Ablation and feature-shuffling studies reveal that the improvements depend on diffusion representation selection, injection positions, and preservation of spatial correspondence. Further evaluation on OmniGen2 demonstrates that DiFuse transfers across unified architectures. These results show that native diffusion representations can serve as complementary spatial cues for multimodal reasoning when selectively integrated in a structure-preserving manner.

In summary, our main contributions are threefold:
\begin{itemize}
\item We investigate direct feature transfer from a UMM’s native diffusion branch to its understanding branch, examining the effects of diffusion state selection, decoder injection position, and spatial correspondence.
\item We propose DiFuse, a post-training framework that aligns intermediate DiT features with visual tokens and incorporates them through residual adapters without an external diffusion or geometry backbone.
\item We demonstrate that DiFuse improves spatially grounded reasoning while largely preserving general multimodal performance on UniWorld-V1, and further validate its transferability on OmniGen2.
\end{itemize}

\begin{figure}[h]
    \centering
    \includegraphics[width=1.0\textwidth]{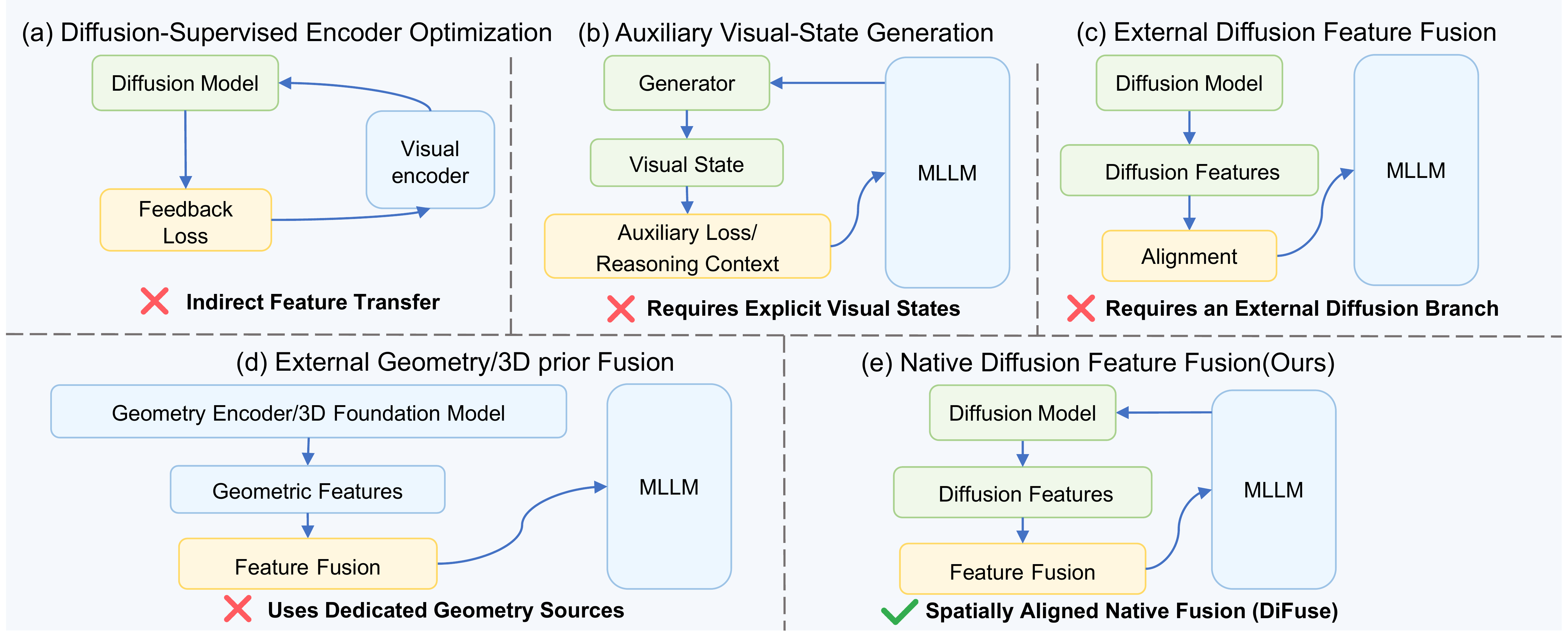}
    \caption{
    \textbf{Paradigms for generation-assisted and structure-enhanced multimodal understanding.} Existing methods use diffusion objectives, explicit visual states, external diffusion features, or dedicated geometry branches, whereas DiFuse directly reuses spatially aligned representations from the UMM's native DiT branch.}
    \label{fig:comparison}
\end{figure}

\section{Related Work}
\subsection{Unified Multimodal Understanding and Generation}
UMMs jointly support visual understanding and image generation within a single architecture~\citep{zhao2025unified}. Existing approaches include token-based autoregressive models such as Chameleon~\citep{team2024chameleon}, Emu3~\citep{wang2024emu3}, and Janus-Pro~\citep{chen2025janus}; diffusion-centric models such as MMaDA~\citep{yang2026mmada} and UniDisc~\citep{swerdlow2025unified}; and hybrid autoregressive-diffusion architectures such as Bagel~\citep{deng2025emerging}, UniWorld-V1~\citep{lin2025uniworld}, and OmniGen2~\citep{wu2025omnigen2}.

Despite these architectural differences, information flow remains largely asymmetric: understanding representations commonly condition generation, whereas intermediate generative representations are rarely reused for understanding. Recent work explores stronger bidirectional interaction through alignment, reconstruction, joint post-training, and benchmark evaluation~\citep{song2026unialignment, xie2026reconstruction, yan2026unified, shi2026realunify, zou2026uni}, but does not directly reuse intermediate representations from a UMM's native diffusion branch as DiFuse does.

\subsection{Generative Representations for Visual Understanding}
Beyond image synthesis, diffusion models learn representations for dense prediction, correspondence, segmentation, and geometric reasoning~\citep{zhao2023unleashing, fuest2026diffusion, gabeur2026image, wang2026diffusion}. As shown in Fig.~\ref{fig:comparison}(a)--(c), existing methods transfer generative or structural information through diffusion feedback and reconstruction objectives, including DIVA~\citep{wang2025diffusion}, DEEM~\citep{luo2025deem}, GenHancer~\citep{ma2025genhancer}, and RecA~\citep{xie2026reconstruction}; explicit visual states, including UniMRG~\citep{su2026generation}, MathCanvas~\citep{shi2026mathcanvas}, Visual Sketchpad~\citep{hu2024visual}, and Latent Sketchpad~\citep{zhang2025latent}; and external diffusion features~\citep{agarwal2025towards, wu2026generation}. In contrast, DiFuse directly reuses intermediate features from the same UMM's native DiT branch, spatially aligns them with visual tokens, and injects them through residual fusion without introducing an additional diffusion or geometry branch.

\subsection{Structural Priors for Spatial Multimodal Reasoning}
Multimodal models often struggle with local spatial relationships and geometric structure~\citep{tong2024eyes, chen2024spatialvlm, wu2026spatial, khayatkhoei2025mllms}, particularly in fine-grained spatial reasoning tasks. Geometry-enhanced MLLMs therefore introduce geometry encoders, 3D foundation models, multiview reconstruction, or explicit geometric supervision~\citep{zheng2026learning, fan2026vlm, hu2026omni, huang20263drs, li2026thinking}.  As illustrated in Fig.~\ref{fig:comparison}(d)--(e), these methods demonstrate the value of structural priors but generally depend on additional geometry sources or supervision, whereas DiFuse reuses native diffusion features within the UMM. Complementary alignment studies suggest that preserving spatial correspondence is important when integrating heterogeneous visual representations~\citep{singh2025matters}, motivating DiFuse to preserve token-level correspondence during residual fusion.

\section{Method}
\label{sec:method}
\begin{figure}[h]
    \centering
    \includegraphics[width=1.0\textwidth]{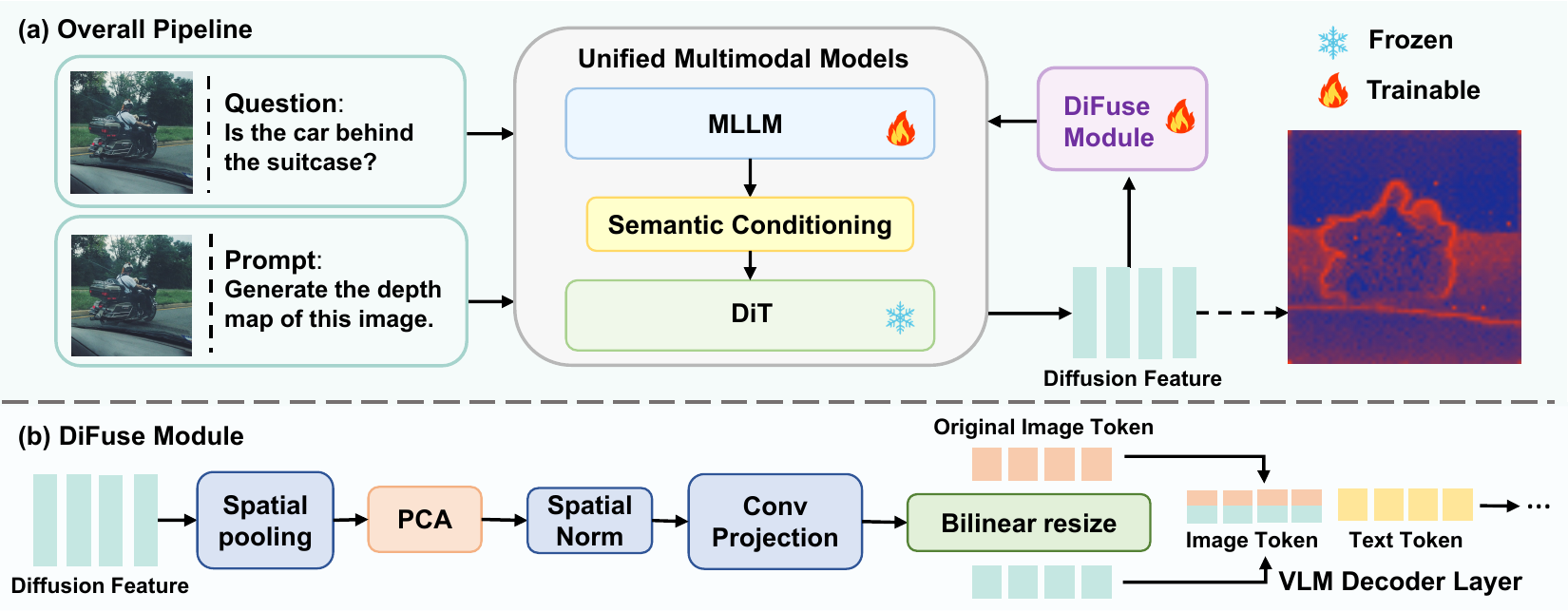}
    \caption{\textbf{DiFuse architecture.}
    \textbf{(a) Overall pipeline.}
    Given an input image and a question, the UMM’s understanding branch processes the visual and text tokens, while its native generation branch receives the same image and a fixed generation prompt “Generate the depth map of this image.” DiFuse extracts an intermediate spatial representation from the frozen DiT at a selected layer and denoising timestep, without decoding a final image.
    \textbf{(b) DiFuse module.}
   The extracted feature is spatially pooled and compressed with PCA, then normalized and projected through a convolutional adapter. Bilinear resizing aligns the resulting feature map with the visual-token grid. DiFuse adds the aligned features to visual-token states at selected decoder layers through residual connections; text-token states receive no direct residual update.}
    \label{fig:architecture}
\end{figure}

In this section, we introduce \textbf{DiFuse}, which incorporates intermediate representations from a UMM's native DiT branch into its understanding pathway. As illustrated in Figure~\ref{fig:architecture}, DiFuse extracts spatial diffusion features, projects them into the language-model hidden space, and injects them into visual-token states at selected decoder layers. The diffusion and visual encoders remain frozen, and training uses the original supervised fine-tuning objective without auxiliary losses.

\subsection{Overview}
\label{sec:method_overview}

Consider a hybrid UMM with an understanding branch $\mathcal{U}$ and a generation branch $\mathcal{G}$. Given an image $x$ and a textual query $q$, the vision encoder and text embedding layer produce a multimodal token sequence
\begin{equation}
    H^{(0)} = [V(x);T(q)] \in \mathbb{R}^{L \times d},
\end{equation}
where $V(x)$ denotes the visual tokens, $T(q)$ denotes the text tokens, and $d$ is the hidden dimension of the multimodal decoder. Let $\mathcal{I}\subseteq\{1,\ldots,L\}$ denote the positions occupied by visual tokens.

In parallel, we obtain an intermediate representation $F^{(k)}_{t}$ from diffusion layer $k$ at denoising timestep $t$. DiFuse transforms this representation with a layer-specific adapter $\mathcal{A}_{\ell}$ and injects it into a selected set of multimodal decoder layers $\mathcal{S}$. If $H^{(\ell)}$ is the output of decoder layer $\ell$, the resulting update is
\begin{equation}
    \widehat{H}^{(\ell)}_{j} =
    \begin{cases}
        H^{(\ell)}_{j} + \lambda_{\ell}\,
        \mathcal{A}_{\ell}\!\left(F^{(k)}_{t}\right)_{\pi(j)},
        & j \in \mathcal{I},\\
        H^{(\ell)}_{j}, & j \notin \mathcal{I},
    \end{cases}
    \qquad \ell\in\mathcal{S},
    \label{eq:difuse_overview}
\end{equation}
where $\pi(j)$ maps a visual-token position to the corresponding location in the aligned diffusion grid, and $\lambda_{\ell}$ is a fixed fusion coefficient. Thus, DiFuse modifies only visual-token states and leaves the text-token pathway intact.

\subsection{Diffusion Feature Extraction}
\label{sec:diffusion_feature_extraction}

\paragraph{Target timestep.}
Given an input image $x$, the native generation branch constructs the diffusion condition $c(x,p_g)$ using the image and a fixed generation prompt $p_g$. Let $\mathcal{T}={t_1,\ldots,t_S}$ denote the timesteps in the inference schedule and let $t^{\star}$ denote the target timestep. Since $t^{\star}$ may not occur exactly in the schedule, we select the closest scheduled timestep:
\begin{equation}
\hat{t}
=
\arg\min_{t_s\in\mathcal{T}}
\left|t_s-t^{\star}\right|.
\label{eq：target_timestep}
\end{equation}
The generation branch follows its native denoising trajectory until $\hat{t}$. We then extract the spatial-token representation from an intermediate DiT layer $k$:
\begin{equation}
R^{(k)}{\hat{t}}
=
\mathcal{G}^{(k)}
\left(z{\hat{t}};c(x,p_g),\hat{t}\right)
\in
\mathbb{R}^{H_DW_D\times d_{\mathrm{raw}}},
\label{eq：diffusion_representation}
\end{equation}
where $z_{\hat{t}}$ is the current noisy latent, $(H_D,W_D)$ denotes the spatial grid of the DiT representation, and $d_{\mathrm{raw}}$ is its channel dimension. We retain the ordered spatial tokens to preserve the spatial correspondence required for location-wise injection. Once extracted, denoising stops without completing the remaining steps or decoding a final image.

\paragraph{Vital layer selection.}
Not all DiT layers provide equally informative representations for multimodal understanding. Following the bypass-based analysis of \citet{avrahami2025stable}, we use layers whose removal substantially alters the generated content as candidate \textit{vital layers}, as these layers are likely to encode influential generative representations. Rather than aggregating features from all candidates, DiFuse selects a single source layer $k^{\star}$ for injection. The selected layer is fixed for each backbone, with the final configuration reported in Section~\ref{sec:experimental_setup} and the corresponding source-layer screening provided in Appendix~\ref{app:ablations}.

\paragraph{Offline caching.}
To avoid repeatedly running the frozen generation branch, we extract the selected diffusion representation once per image and cache it offline. We apply spatial downsampling followed by a fixed PCA projection along the channel dimension,
\begin{equation}
F^{(k^{\star})}_{\hat{t}}
=
\mathcal{P}_{\mathrm{PCA}}
\left(
\mathcal{P}_{s}
\left(
R^{(k^{\star})}_{\hat{t}}
\right)
\right)
\in
\mathbb{R}^{N_D\times d_D}.
\end{equation}
Both operations preserve spatial-token ordering, so the cached representation remains spatially structured. Detailed compression settings are provided in Appendix~\ref{app:feature_cache}.

\subsection{Diffusion Feature Integration and Training}
\label{sec:feature_adaptation}

The cached diffusion feature and the visual tokens generally differ in both channel dimension and spatial resolution. For every injection layer $\ell\in\mathcal{S}$, DiFuse therefore uses a separate adapter $\mathcal{A}{\ell}$.

\paragraph{Structure-preserving projection.}
We first normalize each diffusion channel over its spatial tokens. For $F\in\mathbb{R}^{N_D\times d_D}$, the normalization used in our implementation is
\begin{equation}
    \operatorname{SN}(F){n,c}
    = \frac{F{n,c}-\gamma\mu_c}
    {\sqrt{\sigma_c^2+\epsilon}},
    \label{eq:spatial_norm}
\end{equation}
where $\mu_c$ and $\sigma_c^2$ are computed across the $N_D$ spatial positions, and $\gamma$ is initialized to one and learned during post-training. The normalized sequence is reshaped into its $H_D\times W_D$ grid and processed by a $3\times3$ convolution, GELU, and a $1\times1$ convolution:
\begin{equation}
    P_{\ell}
    = W^{o}{\ell},
    \operatorname{Flat}!\left(
    \operatorname{Conv}^{1\times1}{\ell}
    \left(\operatorname{GELU}\left(
    \operatorname{Conv}^{3\times3}{\ell}
    \left(\operatorname{Grid}(\operatorname{SN}(F))\right)
    \right)\right)\right),
    \label{eq:dit_projection}
\end{equation}
where the convolutions map the diffusion channels to the decoder hidden dimension and $W^{o}{\ell}$ is an output projection. We initialize $W^{o}_{\ell}$ to zero, so the newly introduced pathway begins as an identity-preserving residual and is activated gradually during training.

\paragraph{Spatial alignment.}
Let $(H_V,W_V)$ be the visual-token grid at the multimodal decoder input. We reshape $P_{\ell}$ to a feature map and use bilinear interpolation to match this grid:
\begin{equation}
    \mathcal{A}{\ell}(F)
    = \operatorname{Flat}!\left(
    \operatorname{Resize}{H_V\times W_V}
    \left(\operatorname{Grid}(P_{\ell})\right)
    \right)
    \in\mathbb{R}^{N_V\times d},
    \label{eq:spatial_alignment}
\end{equation}
where $N_V=H_VW_V$. The aligned features are associated with visual tokens in raster order and are written back only at the positions specified by $\mathcal{V}$. This explicit correspondence is important: it transfers local structural cues without altering non-visual tokens.

\paragraph{Fixed residual injection.}
DiFuse uses a fixed scalar $\lambda_{\ell}$ rather than a learned or query-conditioned gate, isolating the contribution of diffusion representations from that of an additional routing network. The injection layers and coefficients are fixed for each backbone, with their settings reported in Section~\ref{sec:experimental_setup} and ablations in Appendix~\ref{app:ablations}.

\paragraph{Training objective.}
DiFuse uses the standard autoregressive SFT objective without auxiliary diffusion, representation-matching, or geometric losses. The original UMM remains frozen, while the DiFuse adapters and decoder LoRA modules are trained with the same data and optimization recipe as the matched SFT baseline. Thus, the two differ only by the diffusion-feature pathway.

\section{Experiments}
\label{sec:experiments}

\subsection{Experimental Setup}
\label{sec:experimental_setup}

\paragraph{Backbones and configurations.}
We conduct our main experiments on UniWorld-V1~\citep{lin2025uniworld} and further evaluate DiFuse on OmniGen2~\citep{wu2025omnigen2} to examine its transferability across hybrid UMM architectures. For UniWorld-V1, our default configuration extracts diffusion representations from layer $g18$ at target timestep $t^{\star}=100$ and injects them into multimodal decoder layers ${0-7}$ using a fixed coefficient $\lambda=1$. Architecture-specific extraction, compression, and injection settings are provided in Appendix~\ref{app:ablations}.

\paragraph{Training.}
We train on 296,401 image--text instruction samples selected from the LLaVA-1.5 mixture~\citep{liu2024improved} after removing GQA-derived samples. For each backbone, DiFuse and its matched SFT baseline use identical training data and optimization settings. We freeze the original model parameters and optimize only the decoder LoRA parameters and DiFuse adapters for three epochs. Complete optimization and hardware configurations are reported in Appendix~\ref{app:implementation}.

\paragraph{Benchmarks.}
We evaluate general multimodal understanding on MMBench~\citep{liu2024mmbench}, MMStar~\citep{chen2024we}, MME~\citep{fu2026mme}, GQA~\citep{hudson2019gqa}, and RealWorldQA~\citep{xai2024realworldqa}. We evaluate spatial understanding on VSR~\citep{liu2023visual}, EmbSpatial~\citep{du2024embspatial}, CV-Bench-3D~\citep{tong2024cambrian}, and 3DSRBench~\citep{ma20253dsrbench}.

\paragraph{Evaluation protocol.}

We follow the official LMMS-Eval protocols~\citep{zhang2024lmmsevalrealitycheckevaluation} and use identical prompting and decoding settings for all variants of the same backbone. We report the normalized MME score (MME$_{\mathrm{norm}}$) as a percentage, placing it on the same scale as the other benchmark scores for fair averaging, and use CircularEval for 3DSRBench (3DSR-C)~\citep{ma20253dsrbench}. Our primary comparison is between DiFuse and its matched SFT baseline, while the original backbone is included as an additional reference. Further evaluation details are provided in Appendix~\ref{app:implementation}.

\subsection{Main Results}

\begin{table*}[t]
\centering
\scriptsize
\setlength{\tabcolsep}{4.0pt}
\renewcommand{\arraystretch}{1.12}

\caption{
\textbf{Controlled comparison of SFT and DiFuse on unified multimodal models.}
For each backbone, SFT and DiFuse use identical instruction-tuning data and optimization settings.
Avg-G averages MME$_{\mathrm{norm}}$, MMB, MMStar, GQA, and RWQA;
Avg-S averages VSR, EmbSpatial, CVB-3D, and 3DSR-C.
Numbers in parentheses denote absolute changes over the corresponding base UMM.
Bold numbers indicate the best result within each backbone family.
}
\label{tab:main_controlled}

\resizebox{\textwidth}{!}{
\begin{tabular}{ll cc ccccc cccc}
\toprule
& &
\multicolumn{2}{c}{\textbf{Summary}} &
\multicolumn{5}{c}{\textbf{General Multimodal Understanding}} &
\multicolumn{4}{c}{\textbf{Spatial Reasoning}} \\
\cmidrule(lr){3-4}
\cmidrule(lr){5-9}
\cmidrule(lr){10-13}

\textbf{Backbone}
& \textbf{Variant}
& \textbf{Avg-G}
& \textbf{Avg-S}
& \textbf{MME$_{\mathrm{norm}}$}
& \textbf{MMB}
& \textbf{MMStar}
& \textbf{GQA}
& \textbf{RWQA}
& \textbf{VSR}
& \textbf{EmbSpatial}
& \textbf{CVB-3D}
& \textbf{3DSR-C} \\
\midrule

\multirow{3}{*}{
    \raisebox{5.0pt}{
    \shortstack[l]{\textit{UniWorld-V1}\\[-1pt]
    \scriptsize (Qwen2.5-VL-7B)}
    }
}
& Base
& 71.92
& 71.08
& \textbf{82.35}
& 83.93
& 62.74
& 60.92
& 69.67
& 87.32
& 69.97
& 81.08
& 45.94 \\

& + SFT
& 71.51 {\scriptsize(-0.41)}
& 71.88 {\scriptsize(+0.80)}
& 78.23
& \textbf{84.88}
& \textbf{63.76}
& 60.90
& \textbf{69.80}
& 87.81
& 70.99
& 81.92
& 46.79 \\

\rowcolor{gray!8}
& \textbf{+ DiFuse}
& \textbf{71.94} {\scriptsize(+0.02)}
& \textbf{73.68} {\scriptsize(+2.60)}
& 80.82
& 84.54
& 62.97
& \textbf{61.70}
& 69.67
& \textbf{89.36}
& \textbf{72.39}
& \textbf{84.58}
& \textbf{48.38} \\

\addlinespace[2pt]
\midrule
\addlinespace[1pt]

\multirow{3}{*}{
    \raisebox{5.0pt}{
    \shortstack[l]{\textit{OmniGen2}\\[-1pt]
    \scriptsize (Qwen2.5-VL-3B)}
    }
}
& Base
& 66.04
& 65.01
& 76.55
& 77.49
& 56.72
& 60.22
& 59.22
& 80.77
& 61.46
& 74.08
& \textbf{43.73} \\

& + SFT
& \textbf{67.69} {\scriptsize(+1.65)}
& 65.37 {\scriptsize(+0.36)}
& \textbf{76.71}
& \textbf{78.26}
& \textbf{58.00}
& \textbf{61.17}
& 64.31
& \textbf{82.41}
& 62.20
& 74.75
& 42.12 \\

\rowcolor{gray!8}
& \textbf{+ DiFuse}
& 67.02 {\scriptsize(+0.98)}
& \textbf{65.84} {\scriptsize(+0.83)}
& 73.15
& 78.01
& 56.44
& 60.96
& \textbf{66.54}
& \textbf{82.41}
& \textbf{63.08}
& \textbf{75.42}
& 42.45 \\

\bottomrule
\end{tabular}
}
\end{table*}

\paragraph{Controlled comparison with matched SFT.}
Table~\ref{tab:main_controlled} presents our primary comparison on UniWorld-V1. The original UMM, the matched SFT baseline, and DiFuse are evaluated under the same protocol, while SFT and DiFuse use identical instruction-tuning data and optimization settings. Therefore, the comparison between SFT and DiFuse isolates the effect of introducing the native diffusion-representation pathway. On UniWorld-V1, DiFuse substantially improves spatial reasoning while preserving overall general multimodal understanding. The spatial average (Avg-S) increases from 71.08 for the original model to 71.88 after standard SFT, and further to 73.68 with DiFuse. This corresponds to a gain of +2.60 points over the original UMM and +1.80 points over the matched SFT baseline.
Importantly, the improvement is consistent across all four primary spatial benchmarks. Compared with SFT, DiFuse improves VSR from 87.81 to 89.36 (+1.55), EmbSpatial from 70.99 to 72.39 (+1.40),
CVB-3D from 81.92 to 84.58 (+2.66), and 3DSR-C from 46.79 to 48.38 (+1.59). These consistent gains indicate that the benefit is not driven by a single spatial benchmark, but transfers across relational, embodied, and 3D spatial reasoning settings.

\paragraph{Preserving general multimodal understanding.}
The effect of DiFuse is considerably more selective on general multimodal benchmarks. Its Avg-G score is 71.94, compared with 71.92 for the original UniWorld-V1 and 71.51 for the SFT baseline. Thus, the spatial improvement does not come at the cost of an overall degradation in general multimodal understanding. At the individual benchmark level, DiFuse improves GQA from 60.90 to 61.70 and MME$_{\mathrm{norm}}$ from 78.23 to 80.82 relative to SFT, while MMB, MMStar, and RWQA remain broadly comparable. This selective behavior is consistent with our motivation, namely diffusion representations primarily provide complementary structural and spatial information rather than a universal enhancement to all multimodal capabilities.

\begin{table*}[t]
\centering
\scriptsize
\setlength{\tabcolsep}{4.0pt}
\renewcommand{\arraystretch}{1.08}

\caption{
\textbf{Contextual comparison with representative general-purpose VLMs, unified multimodal models (UMMs), and spatial-reasoning-enhanced models.}
All models are evaluated using the same benchmark splits and scoring implementation under deterministic decoding.
\textbf{Avg-G} and \textbf{Avg-S} follow the definitions in Table~\ref{tab:main_controlled}.
Bold numbers in the summary columns indicate the best result within each model category.
The shaded row denotes our DiFuse-enhanced UMM.
}
\label{tab:contextual_comparison}

\resizebox{\textwidth}{!}{
\begin{tabular}{l c|cc|ccccc|cccc}
\toprule
& &
\multicolumn{2}{c|}{\textbf{Summary}} &
\multicolumn{5}{c|}{\textbf{General Multimodal Understanding}} &
\multicolumn{4}{c}{\textbf{Spatial Reasoning}} \\

\textbf{Method}
& \textbf{Params.}
& \textbf{Avg-G}
& \textbf{Avg-S}
& \textbf{MME$_{\mathrm{norm}}$}
& \textbf{MMB}
& \textbf{MMStar}
& \textbf{GQA}
& \textbf{RWQA}
& \textbf{VSR}
& \textbf{EmbSpatial}
& \textbf{CVB-3D}
& \textbf{3DSR-C} \\
\midrule

\multicolumn{13}{l}{\textit{General-purpose VLMs}} \\

InternVL3
& 8B
& \textbf{74.24}
& \textbf{72.41}
& 85.18
& 85.74
& 68.65
& 62.73
& 68.89
& 87.07
& 74.45
& 84.00
& 44.11 \\

LLaVA-1.5
& 7B
& 56.51
& 47.16
& 66.89
& 64.00
& 33.76
& 61.93
& 55.95
& 51.47
& 45.77
& 57.00
& 34.38 \\

Llama-3.2-Vision-Instruct
& 11B
& 52.37
& 52.77
& 62.46
& 72.77
& 48.44
& 21.70
& 56.47
& 72.59
& 51.76
& 64.58
& 22.14 \\

Idefics2
& 8B
& 59.46
& 62.50
& 63.99
& 75.26
& 47.34
& 51.22
& 59.48
& 78.56
& 53.49
& 71.00
& 46.93 \\

\midrule

\multicolumn{13}{l}{\textit{Unified Multimodal Models}} \\

BAGEL-MoT
& 7B
& \textbf{74.36}
& \textbf{71.67}
& 83.93
& 84.79
& 69.78
& 66.39
& 66.93
& 81.26
& 70.66
& 84.00
& 50.75 \\

Janus-Pro
& 7B
& 61.14
& 58.08
& 67.03
& 69.24
& 48.63
& 62.35
& 58.43
& 72.67
& 56.62
& 59.50
& 43.51 \\

EMU3-chat
& 8B
& 54.42
& 46.99
& 53.93
& 59.11
& 44.69
& 60.38
& 53.99
& 59.33
& 37.58
& 53.25
& 37.79 \\

\midrule

\multicolumn{13}{l}{
    \textit{Spatial Reasoning Enhanced Models based on Qwen2.5-VL-7B}
} \\

VST-SFT
& 7B
& 70.95
& \textbf{77.17}
& 74.57
& 83.85
& 63.36
& 64.10
& 68.89
& 86.58
& 73.63
& 95.50
& 52.98 \\

GeoThinker
& 7B
& 61.33
& 64.08
& 68.37
& 81.87
& 55.91
& 48.86
& 51.63
& 47.95
& 72.64
& 85.17
& 50.57 \\

ViGoRL-Spatial
& 7B
& 71.31
& 73.75
& 82.62
& 82.22
& 61.95
& 61.13
& 68.63
& 88.95
& 73.43
& 87.42
& 45.22 \\

\rowcolor{gray!8}
\textbf{UniWorld-V1 + DiFuse}
& 7B
& \textbf{71.94}
& 73.68
& 80.82
& 84.54
& 62.97
& 61.70
& 69.67
& 89.36
& 72.39
& 84.58
& 48.38 \\

\bottomrule
\end{tabular}
}
\end{table*}

\paragraph{Comparison with representative multimodal models.}
Table~\ref{tab:contextual_comparison} contextualizes DiFuse among general-purpose VLMs, existing UMMs, and models specifically enhanced for spatial reasoning. UniWorld-V1 + DiFuse achieves an Avg-G of 71.94 and an Avg-S of 73.68. Among the Qwen2.5-VL-7B-based spatially enhanced models, DiFuse retains the strongest general-understanding average while reaching spatial performance comparable to ViGoRL-Spatial (73.68 vs.\ 73.75). Dedicated spatial enhancement methods such as VST-SFT achieve a higher absolute Avg-S, reflecting their explicit optimization for spatial reasoning. In contrast, DiFuse addresses a complementary setting: it reuses the native diffusion representations already available inside a hybrid UMM, without introducing a dedicated geometry encoder or additional geometric supervision. Together with the controlled results above, this comparison suggests that native diffusion representations provide a useful complementary spatial prior for unified multimodal understanding.

Additional qualitative comparisons in Appendix~\ref{app:qualitative} further illustrate that DiFuse primarily corrects errors involving spatial relations, relative depth, and geometric reasoning while largely preserving predictions on general visual understanding tasks.

\subsection{Ablation Studies}
\label{sec:ablations}
We conduct ablation studies to examine how the diffusion timestep, VLM decoder injection layers, fusion coefficient, and spatial correspondence affect DiFuse. Unless otherwise specified, experiments are conducted on UniWorld-V1 using LoRA tuning for the Qwen2.5-VL backbone, while the native DiT remains frozen. We use the Avg-G and Avg-S metrics defined in Table~\ref{tab:main_controlled}.

\paragraph{Diffusion timestep.}
As shown in Table~\ref{tab:ablation_config}, timestep 100 achieves the best overall balance between general and spatial performance, while both lower and higher timesteps are less consistent across benchmarks. We therefore use $t=100$ in the final configuration. Timestep-wise feature visualizations are provided in the Appendix~\ref{app:representation_analysis}

\paragraph{VLM decoder injection layers.}
Prior analysis shows that multimodal LLMs depend more strongly on non-textual tokens in earlier layers, with this dependence weakening as cross-modal fusion progresses~\citep{xin2026stage}. Accordingly, for the 28-layer Qwen2.5-VL decoder, we use layer 7 as a quarter-depth anchor and evaluate injection over approximately the first quarter (layers 0--7), the first half (layers 0--13), and narrower early-layer ranges. Injecting into layers 0--7 achieves the best balance, with an Avg-G of 71.94 and an Avg-S of 73.68. Narrower ranges provide weaker spatial gains, while extending injection to layers 0--13 does not improve spatial performance and reduces general capability. This suggests that diffusion features are most useful during early cross-modal integration within the understanding pathway.

\paragraph{Diffusion source-layer selection.}
Based by the vital-layer analysis of Stable Flow~\citep{avrahami2025stable}, we screen $g17$, $g18$, $g28$, and $g53$, covering the dual-to-single-stream boundary and later single-stream regions. All candidates are evaluated under the same reduced configuration to isolate the effect of the source layer. As Table~\ref{tab:source_layer_main} shows that the performance does not vary monotonically with diffusion depth: $g28$ achieves the highest Avg-S of 73.18, but exceeds $g18$ by only 0.10 points, whereas $g18$ achieves the highest Avg-G of 71.89 and leads $g28$ by 0.51 points. We therefore select $g18$ for its better general--spatial balance. These screening results are not directly comparable to the final DiFuse configuration; further details are provided in Appendix~\ref{app:source_layer_screening}.

\begin{table*}[t]
\centering
\scriptsize
\setlength{\tabcolsep}{4.0pt}
\renewcommand{\arraystretch}{1.05}

\caption{\textbf{Diffusion source-layer screening.}
All candidates are evaluated under the same reduced screening configuration with all other settings fixed.
These values are not directly comparable to the final DiFuse results.
Bold numbers indicate the best result in each column, and the shaded row denotes the selected source layer.}
\label{tab:source_layer_main}

\resizebox{\textwidth}{!}{
\begin{tabular}{l|cc|ccccc|cccc}
\toprule
&
\multicolumn{2}{c|}{\textbf{Summary}} &
\multicolumn{5}{c|}{\textbf{General Multimodal Understanding}} &
\multicolumn{4}{c}{\textbf{Spatial Reasoning}} \\

\textbf{Source layer}
& \textbf{Avg-G}
& \textbf{Avg-S}
& \textbf{MME$_{\mathrm{norm}}$}
& \textbf{MMB}
& \textbf{MMStar}
& \textbf{GQA}
& \textbf{RWQA}
& \textbf{VSR}
& \textbf{EmbSpatial}
& \textbf{CVB-3D}
& \textbf{3DSR-C} \\
\midrule

$g17$
& 71.44
& 72.14
& 79.00
& 83.76
& 62.61
& 61.11
& 70.72
& 88.30
& 71.18
& 81.58
& 47.51 \\

\rowcolor{gray!8}
$\mathbf{g18}$
& \textbf{71.89}
& 73.08
& 81.31
& 83.51
& 64.03
& 60.92
& 69.67
& 89.28
& 72.12
& 82.83
& 48.09 \\

$g28$
& 71.38
& \textbf{73.18}
& 79.95
& 83.93
& 62.39
& 61.23
& 69.41
& 88.87
& 72.17
& 83.67
& 48.03 \\

$g53$
& 71.74
& 73.06
& 79.95
& 84.36
& 62.46
& 60.92
& 70.98
& 89.12
& 71.95
& 83.00
& 48.19 \\

\bottomrule
\end{tabular}
}
\end{table*}

\paragraph{Fusion coefficient.}
We vary the fusion coefficient $\lambda$ over $\{0.25,0.5,0.75,1.0,1.25\}$ to examine the effect of diffusion-feature injection strength. As shown in Table~\ref{tab:ablation_config}, different injection strengths lead to different trade-offs between general and spatial performance. Among the evaluated settings, $\lambda = 1.0$ achieves the highest Avg-S of 73.68 while maintaining an Avg-G of 71.94, close to the best general-performance result. Increasing $\lambda$ further to 1.25 reduces both Avg-G and Avg-S. We therefore adopt $\lambda$ as the final setting based on its overall balance between spatial reasoning and general multimodal performance.

\paragraph{Spatial correspondence.}
Table~\ref{tab:ablation_spatial} examines whether DiFuse benefits from diffusion features themselves and whether their spatial correspondence with visual tokens matters. Image shuffling replaces the diffusion features with those from another image, reducing Avg-S from 73.68 to 72.60 and highlighting the importance of image-specific information. Token shuffling preserves features from the correct image but permutes their spatial positions, causing the largest drop to 71.84 and demonstrating that correct content alone is insufficient without spatial correspondence. Random features achieve an Avg-S of 72.80, suggesting that the additional residual pathway itself may contribute to part of the improvement. Nevertheless, spatially aligned diffusion features further improve Avg-S to 73.68, indicating that structured diffusion representations provide benefits beyond additional fusion capacity. Avg-G remains relatively stable across these controls, suggesting that the observed effect is primarily concentrated on spatial reasoning.

\begin{table*}[t]
\centering
\scriptsize
\setlength{\tabcolsep}{4.0pt}
\renewcommand{\arraystretch}{1.08}

\caption{
\textbf{Ablation of core DiFuse design choices on UniWorld-V1.}
Each block varies one factor while fixing the others; shaded entries denote the final configuration.}
\label{tab:ablation_config}

\resizebox{\textwidth}{!}{
\begin{tabular}{l c|cc|ccccc|cccc}
\toprule
& &
\multicolumn{2}{c|}{\textbf{Summary}} &
\multicolumn{5}{c|}{\textbf{General Multimodal Understanding}} &
\multicolumn{4}{c}{\textbf{Spatial Reasoning}} \\

\textbf{Factor}
& \textbf{Variant}
& \textbf{Avg-G}
& \textbf{Avg-S}
& \textbf{MME$_{\mathrm{norm}}$}
& \textbf{MMB}
& \textbf{MMStar}
& \textbf{GQA}
& \textbf{RWQA}
& \textbf{VSR}
& \textbf{EmbSpatial}
& \textbf{CVB-3D}
& \textbf{3DSR-C} \\
\midrule

\multicolumn{13}{l}{\textit{Diffusion timestep}} \\

$t$
& 50
& 71.52
& 72.98
& 79.84
& 83.76
& 62.87
& 61.48
& 69.67
& 88.95
& 71.76
& 83.58
& 47.64 \\

\rowcolor{gray!8}
$t$
& \textbf{100 (ours)}
& \textbf{71.94}
& \textbf{73.68}
& 80.82
& 84.54
& 62.97
& 61.70
& 69.67
& 89.36
& 72.39
& 84.58
& 48.38 \\

$t$
& 200
& 71.59
& 73.05
& 80.33
& 83.93
& 62.53
& 61.32
& 69.80
& 89.77
& 72.39
& 81.50
& 48.54 \\

$t$
& 300
& 71.50
& 73.52
& 80.29
& 84.19
& 62.83
& 61.43
& 68.76
& 89.12
& 72.31
& 83.75
& 48.90 \\

\midrule

\multicolumn{13}{l}{\textit{VLM decoder injection layers}} \\

\text{Decoder layers}
& 7
& 71.63
& 73.15
& 80.58
& 83.85
& 62.59
& 60.30
& 70.85
& 89.53
& 70.30
& 83.25
& 49.52 \\

\text{Decoder layers}
& 0--3
& 71.89
& 73.08
& 81.31
& 83.51
& 64.03
& 60.92
& 69.67
& 89.28
& 72.12
& 82.83
& 48.09 \\

\rowcolor{gray!8}
\text{Decoder layers}
& \textbf{0--7 (ours)}
& \textbf{71.94}
& \textbf{73.68}
& 80.82
& 84.54
& 62.97
& 61.70
& 69.67
& 89.36
& 72.39
& 84.58
& 48.38 \\

\text{Decoder layers}
& 4--7
& 71.65
& 73.05
& 80.28
& 84.28
& 63.46
& 61.49
& 68.76
& 88.87
& 72.45
& 83.00
& 47.90 \\

\text{Decoder layers}
& 0--13
& 71.11
& 73.08
& 79.35
& 84.36
& 62.13
& 61.07
& 68.63
& 89.12
& 71.62
& 83.67
& 47.92 \\

\midrule

\multicolumn{13}{l}{\textit{Fusion coefficient}} \\

$\lambda$
& 0.25
& 71.74
& 72.88
& 81.51
& 83.93
& 62.64
& 61.35
& 69.28
& 88.87
& 71.65
& 82.75
& 48.26 \\

$\lambda$
& 0.50
& \textbf{72.00}
& 72.08
& 81.44
& 84.45
& 63.22
& 60.04
& 70.85
& 88.30
& 69.81
& 81.17
& 49.06 \\

$\lambda$
& 0.75
& 71.84
& 73.42
& 80.09
& 84.19
& 63.64
& 61.48
& 69.80
& 89.53
& 72.47
& 83.83
& 47.86 \\

\rowcolor{gray!8}
$\lambda$
& \textbf{1.00 (ours)}
& 71.94
& \textbf{73.68}
& 80.82
& 84.54
& 62.97
& 61.70
& 69.67
& 89.36
& 72.39
& 84.58
& 48.38 \\

$\lambda$
& 1.25
& 71.40
& 73.29
& 81.41
& 83.51
& 62.37
& 61.60
& 68.10
& 89.03
& 72.66
& 83.83
& 47.64 \\

\bottomrule
\end{tabular}
}
\end{table*}

\begin{table*}[t]
\centering
\scriptsize
\setlength{\tabcolsep}{4.0pt}
\renewcommand{\arraystretch}{1.08}
\caption{
\textbf{Effect of spatial correspondence in DiFuse.}
Image and token shuffling disrupt the corresponding image-level and token-level spatial alignment, while random features replace the diffusion representation. The shaded row denotes the final configuration.
}
\label{tab:ablation_spatial}

\resizebox{\textwidth}{!}{
\begin{tabular}{l c|cc|ccccc|cccc}
\toprule
& &
\multicolumn{2}{c|}{\textbf{Summary}} &
\multicolumn{5}{c|}{\textbf{General Multimodal Understanding}} &
\multicolumn{4}{c}{\textbf{Spatial Reasoning}} \\

\textbf{Condition}
& \textbf{Variant}
& \textbf{Avg-G}
& \textbf{Avg-S}
& \textbf{MME$_{\mathrm{norm}}$}
& \textbf{MMB}
& \textbf{MMStar}
& \textbf{GQA}
& \textbf{RWQA}
& \textbf{VSR}
& \textbf{EmbSpatial}
& \textbf{CVB-3D}
& \textbf{3DSR-C} \\
\midrule

\rowcolor{gray!8}
\text{Correspondence}
& \textbf{Aligned features (ours)}
& \textbf{71.94}
& \textbf{73.68}
& 80.82
& 84.54
& 62.97
& 61.70
& 69.67
& 89.36
& 72.39
& 84.58
& 48.38 \\

\text{Correspondence}
& Image-shuffled features
& 71.33
& 72.60
& 80.25
& 84.19
& 62.52
& 61.46
& 68.24
& 88.87
& 71.65
& 82.83
& 47.04 \\

\text{Correspondence}
& Token-shuffled features
& 71.80
& 71.84
& 81.59
& 83.93
& 62.41
& 61.14
& 69.93
& 87.81
& 70.80
& 81.42
& 47.33 \\

\text{Correspondence}
& Random features
& 71.91
& 72.80
& 80.36
& 84.45
& 64.73
& 60.74
& 69.28
& 88.95
& 72.01
& 82.00
& 48.23 \\

\bottomrule
\end{tabular}
}
\end{table*}

\section{Conclusion}

We investigate whether a UMM's native diffusion branch can enhance understanding, particularly spatial reasoning. To this end, we introduce DiFuse, which reuses intermediate DiT representations from the generative branch, spatially aligns them with visual tokens, and injects them into selected decoder layers of the understanding branch. On UniWorld-V1, DiFuse improves Avg-S by 1.80 points over the matched SFT baseline while largely preserving general multimodal performance, and results on OmniGen2 further support its applicability across different UMM architectures. Our analyses show that these gains depend on selecting informative diffusion states and preserving their spatial correspondence. Together, these findings suggest that native diffusion representations provide UMMs with complementary spatial and structural cues, establishing a practical generation-to-understanding pathway for spatial reasoning. Future work may include adaptive feature selection, broader UMM architectures, and joint evaluation of understanding and generation.

\newpage
\section*{AI use statement}

In this work, we used generative AI tools for the following tasks:
design or provide feedback on research methodology or experiments;
implement methods;
propose or refine hypotheses;
assist with translation;
support qualitative and thematic data analysis;
and interpret results.
We have not used generative AI tools for the following tasks:
Generate synthetic data sets;
help develop theoretical models or conceptual frameworks;
and clean and reformat dataset.
The following tasks are not applicable to this work:
formulate mathematical claims;
provide critical ingredients for proving mathematical claims;
and assist in the writing of proofs.

Additionally, we used generative AI tools to
suggest experimental parameters;
create or edit software code;
draft parts of a research paper;
summarize or analyse existing literature;
discover research topics or identify gaps;
sourcing/searching for information;
edit a research paper to improve readability;
format references;
and suggest a structure for a research paper.

We take responsibility for the final content of this work, including text, claims, code, figures, and other artifacts produced with the aid of generative AI.

\section*{Ethics Statement}

This work studies multimodal understanding and image generation using publicly available models and benchmark datasets. It does not involve human-subject experiments, the collection of private personal data, or biological or chemical applications. Model outputs may inherit biases or errors from the underlying models and datasets. The proposed system should not be used in safety-critical or high-stakes decision-making without additional reliability, fairness, and privacy auditing.

\subsection*{Reproducibility Statement}

We will release code, configurations, checkpoints, feature-extraction utilities, and evaluation scripts. Section~\ref{sec:method} describes the DiFuse architecture and training procedure, while Section~\ref{sec:experimental_setup} specifies the model configurations, datasets, benchmarks, metrics, and evaluation protocols. Section~\ref{sec:ablations} reports the main design ablations, and the appendix provides implementation details, extended benchmark results, additional ablations, visualizations, and qualitative examples. All evaluations use fixed benchmark splits, consistent scoring protocols, official model templates, and deterministic decoding.

\bibliography{iclr2027_conference}
\bibliographystyle{iclr2027_conference}

\newpage
\appendix
\section{Additional Implementation Details}
\label{app:implementation}

This section provides backbone-specific configurations, optimization hyperparameters, and model-side computational costs that are omitted from the main paper. The common DiFuse operations, including feature extraction, spatial alignment, and residual injection, are defined in Section~\ref{sec:method} and are not repeated here.

\subsection{Model and Fusion Configuration}
\label{app:model_configuration}

\begin{table}[h]
    \centering
    \scriptsize
    \setlength{\tabcolsep}{4.2pt}
    \renewcommand{\arraystretch}{1.05}

    \caption{
    \textbf{Backbone-specific DiFuse configurations.}
    DiT layer denotes the source layer in the native generation branch;
    injection layers denote layers in the multimodal decoder.
    }
    \label{tab:app_model_configuration}

    \begin{tabular}{@{}lcc@{}}
        \toprule
        \textbf{Configuration}
        & \textbf{UniWorld-V1}
        & \textbf{OmniGen2} \\
        \midrule
        Understanding backbone
        & Qwen2.5-VL-7B
        & Qwen2.5-VL-3B \\

        Decoder hidden dimension
        & 3,584
        & 2,048 \\

        Decoder layers
        & 28
        & 36 \\

        DiT source layer
        & 18
        & 6 \\

        Target timestep
        & 100
        & 900 \\

        Cached feature dimension
        & 1,024
        & 1,024 \\

        Injection layers
        & 0--7
        & 0--9 \\

        Number of fusion adapters
        & 8
        & 10 \\

        Fixed fusion coefficient
        & 1.0
        & 0.5 \\

        LoRA-enabled decoder layers
        & 0--27
        & 0--35 \\

        \bottomrule
    \end{tabular}
\end{table}

Table~\ref{tab:app_model_configuration} summarizes the architecture-specific settings used for UniWorld-V1 and OmniGen2. Both implementations follow the same fusion principle, while using backbone-specific DiT source layers, diffusion timesteps, and fusion coefficients. Since the two understanding branches have different hidden dimensions, their fusion adapters are trained independently, and no projection weights are transferred between the two backbones.

To obtain a consistent and interpretable configuration across backbones, DiFuse injects diffusion representations into a shallow prefix of the multimodal decoder. Specifically, we use approximately the first quarter of decoder layers: layers 0--7 of the 28-layer decoder in UniWorld-V1 and layers 0--9 of the 36-layer decoder in OmniGen2. This proportional selection keeps the relative injection depth comparable across architectures while preserving the early semantic processing pathway.

\subsection{Training and Evaluation Protocols}
\label{app:training_protocol}

\begin{table}[h]
    \centering
    \scriptsize
    \setlength{\tabcolsep}{4.2pt}
    \renewcommand{\arraystretch}{1.05}

    \caption{
    \textbf{Training configuration for the UniWorld-V1 experiments.}
    The effective batch size includes gradient accumulation over all GPUs.
    }
    \label{tab:app_training_configuration}

    \begin{tabular}{@{}lc@{}}
        \toprule
        \textbf{Setting} & \textbf{Value} \\
        \midrule
        Training examples & 296,401 \\
        Training epochs & 3 \\
        Optimization steps & 6,948 \\
        Hardware & $8\times$ NVIDIA A100-80GB \\
        Per-device batch size & 2 \\
        Gradient accumulation & 8 \\
        Effective global batch size & 128 \\
        Numerical precision & BF16 \\
        Distributed optimization & DeepSpeed ZeRO-2 \\
        Optimizer & AdamW \\
        AdamW $(\beta_1,\beta_2)$ & $(0.9,0.999)$ \\
        Weight decay & 0.01 \\
        Maximum gradient norm & 1.0 \\
        Learning-rate schedule & Cosine decay \\
        Warm-up steps & 1,000 \\
        LoRA learning rate & $1\times10^{-6}$ \\
        Fusion-adapter learning rate & $5\times10^{-5}$ \\
        LoRA rank / scaling & 128 / 256 \\
        LoRA dropout & 0.05 \\
        \bottomrule
    \end{tabular}
\end{table}

Table~\ref{tab:app_training_configuration} lists the optimization settings used in our main UniWorld-V1 experiments. We train the matched SFT and DiFuse models for the same number of optimization steps using the same instruction data and LoRA configuration. The original vision encoder, DiT, VAE, and multimodal decoder weights remain frozen; only the LoRA parameters and DiFuse adapters are optimized.

LoRA is applied to the query, key, value, and output projections in self-attention and to the gate, up, and down projections in the decoder MLP. For OmniGen2, the SFT and DiFuse variants are also trained for three epochs and 6,948 optimization steps on the same approximately 296.4K-example subset. Within each backbone, SFT and DiFuse use the same effective batch size, training schedule, and LoRA configuration.

All model variants are evaluated using the same LMMS-Eval task definitions, splits, and scoring. We retain each model's official chat template and use its recommended reasoning prompt when required. Evaluation uses non-sampling decoding with temperature zero and one beam. The maximum output length follows the task configuration: for example, MME uses at most 16 generated tokens, while tasks without an explicit override use the evaluator default of 1,024 tokens.

\subsection{Trainable Parameters and Storage Costs}
\label{app:computational_cost}

\begin{table}[h]
    \centering
    \scriptsize
    \setlength{\tabcolsep}{4.2pt}
    \renewcommand{\arraystretch}{1.05}

    \caption{
    \textbf{Trainable parameter and checkpoint storage costs for the final DiFuse configurations.}
    The reported values correspond to the main UniWorld-V1 experiment and the OmniGen2 generalization experiment. Parameter counts may vary across ablation settings that change the number of injection layers or the feature dimension.
    }
    \label{tab:app_model_cost}

    \begin{tabular}{@{}l*{4}{c}@{}}
        \toprule
        \textbf{Backbone}
        & \textbf{LoRA Params.}
        & \textbf{Fusion Params.}
        & \textbf{LoRA Size}
        & \textbf{Fusion Size} \\
        \midrule
        UniWorld-V1
        & 323.0M
        & 207.7M
        & 646.0 MB
        & 415.3 MB \\

        OmniGen2
        & 239.5M
        & 157.3M
        & 957.9 MB
        & 314.6 MB \\
        \bottomrule
    \end{tabular}
\end{table}

Table~\ref{tab:app_model_cost} reports costs that can be determined directly from the trained checkpoints. Since the original UMM parameters are frozen, the trainable parameters consist only of LoRA weights and fusion adapters. For UniWorld-V1, these components contain approximately 530.6M parameters in total, corresponding to about $2.5\%$ of the complete 20.90B-parameter system loaded during training.

Offline extraction removes repeated DiT computation from instruction tuning: each training image is processed by the frozen generation branch once, and the resulting feature is reused across epochs. The cache size depends on the number of spatial tokens associated with each image and cannot be inferred from the model checkpoint alone. We therefore report model-side storage independently of dataset-dependent cache storage.

\subsection{Feature Compression and Offline Caching}
\label{app:feature_cache}

To avoid repeatedly running the frozen generation branch during training and evaluation, we extract the selected diffusion representation once for each image and store it offline. For both UniWorld-V1 and OmniGen2, the extracted spatial feature grid at 1,024-pixel resolution is reduced by adaptive average pooling to the grid associated with a 512-pixel anchor. We then apply a fixed PCA projection independently to each spatial token, reducing the channel dimension from $3{,}072$ to $1{,}024$. The PCA basis is fitted once and kept fixed, with the retained components explaining more than $99.5\%$ of the cumulative variance in the fitting features.

These operations substantially reduce feature-storage, data-loading, and adaptation costs while preserving the two-dimensional organization of the diffusion representation. In particular, spatial pooling reduces the number of spatial tokens without collapsing them into a global descriptor, while PCA operates only along the channel dimension. Therefore, neither operation changes the ordering of the tokens, allowing the cached features to retain the spatial correspondence required by DiFuse.

\section{Detailed Benchmark Results}
\label{app:detailed_results}

To diagnose the spatial improvements in more detail, we report category-level results on EmbSpatial and CV-Bench-3D. As Table~\ref{tab:app_spatial_breakdown} shows, EmbSpatial is divided into AI2-THOR, Matterport3D, and ScanNet, while CV-Bench-3D contains two subtasks: Depth Order and Relative Distance. VSR and 3DSR-C are reported only using their aggregate scores because their category-level results are either unavailable or not directly comparable under the adopted evaluation protocols.

\begin{table*}[h]
    \centering
    \scriptsize
    \setlength{\tabcolsep}{4.2pt}
    \renewcommand{\arraystretch}{1.05}

    \caption{
    \textbf{Category-wise spatial reasoning results on EmbSpatial and CV-Bench-3D (\%).}
    EmbSpatial is divided into AI2-THOR, Matterport3D, and ScanNet.
    CV-Bench-3D reports Depth Order and Relative Distance subtasks.
    The Overall column denotes the official aggregate accuracy.
    }

    \label{tab:app_spatial_breakdown}
    \begin{tabular}{ll|cccc|ccc}
        \toprule
        & &
        \multicolumn{4}{c|}{\textbf{EmbSpatial}} &
        \multicolumn{3}{c}{\textbf{CV-Bench-3D}} \\

        \textbf{Backbone}
        & \textbf{Variant}
        & \textbf{AI2-THOR}
        & \textbf{MP3D}
        & \textbf{ScanNet}
        & \textbf{Overall}
        & \textbf{Depth}
        & \textbf{Distance}
        & \textbf{Overall} \\
        \midrule

        & Base
        & 68.93 & 71.61 & 69.42 & 69.97
        & 83.33 & 78.83 & 81.08 \\

        UniWorld-V1
        & + SFT
        & 70.86 & 72.11 & 70.00 & 70.99
        & 83.17 & 80.67 & 81.92 \\

        \rowcolor{gray!8}
        & + DiFuse
        & 72.64 & 73.36 & 71.17 & 72.39
        & 85.33 & 83.83 & 84.58 \\

        \midrule

        & Base
        & 55.45 & 65.11 & 64.00 & 61.46
        & 76.50 & 71.67 & 74.08 \\

        OmniGen2
        & + SFT
        & 56.74 & 64.61 & 65.42 & 62.20
        & 79.50 & 70.00 & 74.75 \\

        \rowcolor{gray!8}
        & + DiFuse
        & 58.51 & 65.03 & 65.83 & 63.08
        & 81.33 & 69.50 & 75.42 \\

        \bottomrule
    \end{tabular}
\end{table*}

For UniWorld-V1, DiFuse improves all three EmbSpatial categories over the base model, with the largest gain on AI2-THOR. It also improves both CV-Bench-3D subtasks, increasing Depth Order from 83.33\% to 85.33\% and Relative Distance from 78.83\% to 83.83\%. Compared with the SFT baseline, DiFuse still achieves consistent gains across both EmbSpatial and CV-Bench-3D.

On OmniGen2, DiFuse improves the EmbSpatial overall score and the Depth Order subtask, while Relative Distance changes only slightly. This result supports the transferability of DiFuse across architectures, although the gains are not uniform across all subcategories.

\section{Extended Ablation Studies}
\label{app:ablations}

We supplement the main ablations by examining the selection of the diffusion source layer and whether combining multiple diffusion representations provides a reliable advantage.

\subsection{Source-Layer Screening}
\label{app:source_layer_screening}

Stable Flow~\citep{avrahami2025stable} identifies ten vital layers in FLUX.1-dev~\citep{flux2024}, $\{g0,g1,g2,g17,g18,\allowbreak
g25,g28,g53,g54,g56\}$, by measuring the perceptual change caused by bypassing individual transformer blocks. Under our global indexing convention, FLUX contains 19 dual-stream blocks ($g0$--$g18$) followed by 38 single-stream blocks ($g19$--$g56$). These layers therefore cover early dual-stream states, the dual-to-single-stream boundary, intermediate single-stream states, and late single-stream states.

We exclude $g0$--$g2$ because they are closest to the noisy latent, and retain $g18$, $g28$, and $g53$ as representative layers from the boundary, intermediate, and late regions. We additionally include $g17$ to test sensitivity to the precise layer choice near the dual-to-single-stream boundary. All candidates use the same reduced screening configuration, with the timestep, feature-processing pipeline, decoder injection range, and fusion coefficient fixed. The corresponding performance results are reported in Table~\ref{tab:source_layer_main}.

\subsection{Exploratory Multi-Layer Feature Fusion}
\label{app:multilayer_fusion}

Based on the source-layer screening, we use $g18$ as the anchor and examine whether an additional diffusion representation provides complementary information. We exclude $g17$ because it is adjacent to $g18$ and performs less favorably in the single-layer screening, and instead consider the non-adjacent $g28$ and $g53$ layers from later regions of the diffusion backbone.

For an auxiliary layer $k\in\{28,53\}$, we map its representation into the $g18$ feature space and add it as a bounded residual:
\begin{equation}
\widetilde{F}^{(\ell)}
=
F_{18}
+
\beta_{\ell,k}\mathcal{A}_{k}(F_k),
\qquad
0 < \beta_{\ell,k} < \beta_{\max},
\end{equation}
where $\mathcal{A}_{k}$ is a learnable linear aligner and $\beta_{\ell,k}$ is learned independently for each VLM injection layer. For all multi-layer variants, $\beta_{\ell,k}$ is initialized to $0.10$ and bounded by $\beta_{\max}=0.30$. Table~\ref{tab:app_multilayer} compares $g28$ and $g53$ under different decoder injection ranges and main fusion coefficients $\lambda$, with a matched single-layer $g18$ reference in each group.

\begin{table*}[h]
\centering
\scriptsize
\setlength{\tabcolsep}{4.2pt}
\renewcommand{\arraystretch}{1.05}

\caption{\textbf{Exploratory anchor-residual multi-layer fusion.}
All multi-layer variants use $g18$ as the anchor and add either $g28$ or $g53$ through a learnable bounded residual.
Rows are grouped by decoder injection range and main fusion coefficient $\lambda$.
Shaded rows denote the matched single-layer references, with the darker row indicating the final DiFuse configuration.
Bold numbers indicate the best Avg-G and Avg-S within each group.}
\label{tab:app_multilayer}

\resizebox{\textwidth}{!}{
\begin{tabular}{l c c|cc|cccc}
\toprule
\textbf{Source layers}
& \textbf{Inject.}
& $\boldsymbol{\lambda}$
& \textbf{Avg-G}
& \textbf{Avg-S}
& \textbf{VSR}
& \textbf{EmbSpatial}
& \textbf{CVB-3D}
& \textbf{3DSR-C} \\
\midrule

\rowcolor{gray!8}
\textbf{Single-layer $g18$}
& $0$--$3$ & 0.5
& 71.89 & 73.08
& 89.28 & 72.12 & 82.83 & 48.09 \\

$g18+g53$
& $0$--$3$ & 0.5
& \textbf{71.90} & 73.04
& 88.87 & 71.40 & 83.75 & 48.15 \\

$g18+g28$
& $0$--$3$ & 0.5
& 71.63 & \textbf{73.23}
& 89.12 & 71.59 & 84.08 & 48.13 \\

\midrule

\rowcolor{gray!8}
\textbf{Single-layer $g18$}
& $0$--$7$ & 0.5
& 72.00 & 72.08
& 88.30 & 69.81 & 81.17 & 49.06 \\

$g18+g28$
& $0$--$7$ & 0.5
& 71.51 & \textbf{73.40}
& 89.12 & 72.34 & 83.75 & 48.38 \\

$g18+g53$
& $0$--$7$ & 0.5
& \textbf{72.02} & 73.00
& 88.63 & 72.25 & 83.17 & 47.97 \\

\midrule

\rowcolor{gray!15}
\textbf{Single-layer $g18$}
& $0$--$7$ & 1.0
& \textbf{71.94} & \textbf{73.68}
& 89.36 & 72.39 & 84.58 & 48.38 \\

$g18+g53$
& $0$--$7$ & 1.0
& 71.66 & 73.54
& 89.12 & 72.20 & 84.42 & 48.42 \\

$g18+g28$
& $0$--$7$ & 1.0
& 71.57 & 73.15
& 89.12 & 72.23 & 83.33 & 47.92 \\

\bottomrule
\end{tabular}
}
\end{table*}

With $\lambda=0.5$, the auxiliary layers provide setting-dependent spatial gains. For injection into layers 0--3, adding $g28$ improves Avg-S from 73.08 to 73.23 at the cost of Avg-G, while adding $g53$ leaves the summary performance nearly unchanged. With injection into layers 0--7, $g28$ achieves the highest Avg-S of 73.40, whereas $g53$ improves Avg-S from 72.08 to 73.00 while preserving Avg-G.

Keeping the same injection range and auxiliary-coefficient setting, increasing $\lambda$ to $1.0$ removes these gains. Relative to the matched single-layer configuration, $g18+g53$ reduces Avg-G and Avg-S by $0.28$ and $0.14$, while $g18+g28$ reduces them by $0.37$ and $0.53$, respectively. The benefit of an auxiliary source therefore depends on its interaction with the main fusion strength and injection range. Since none of the explored multi-layer configurations consistently improves over its matched single-layer reference, we retain the simpler single-layer $g18$ design.

\section{Visualization of Diffusion Representations}
\label{app:representation_analysis}

\begin{figure}[h]
    \centering
    \includegraphics[width=\linewidth]{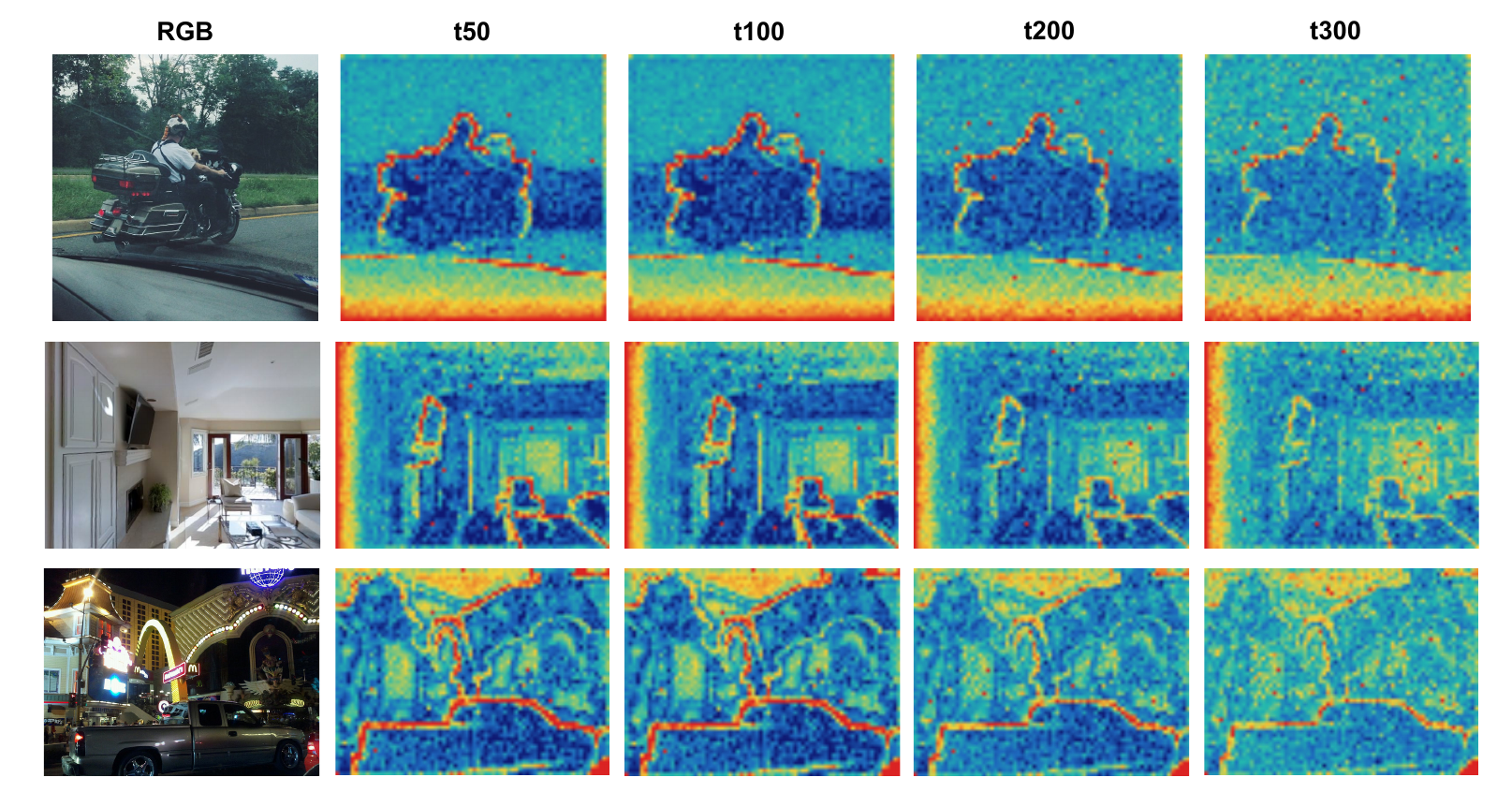}
    \caption{\textbf{Diffusion responses across timesteps at fixed layer $g18$.} Response maps are shown at $t\in\{50,100,200,300\}$ for three representative images. Spatial organization is retained across timesteps, while the response granularity varies.}
    \label{fig:timestep_representation}
\end{figure}

\begin{figure}[h]
    \centering
    \includegraphics[width=\linewidth]{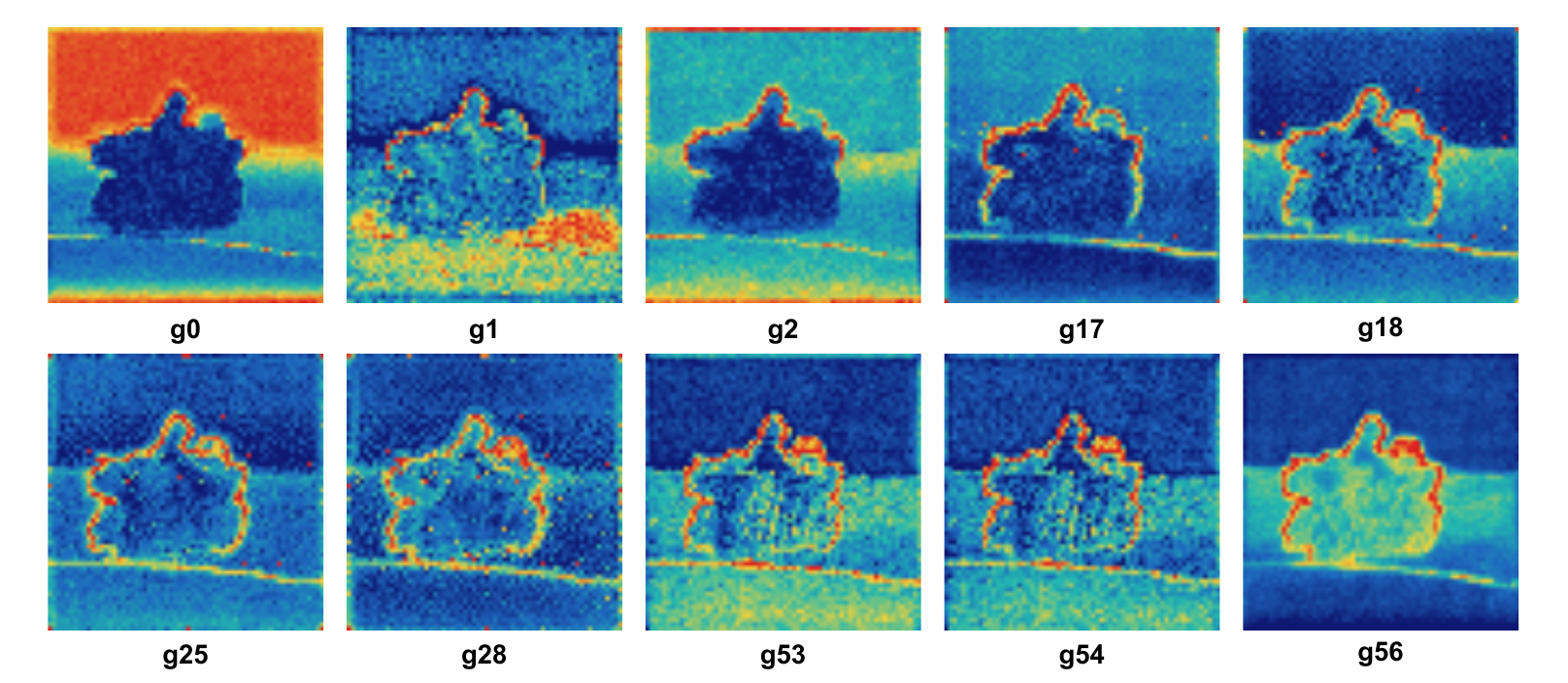}
    \caption{
    \textbf{Diffusion responses across candidate layers at fixed timestep $t=100$.} Response maps are shown for the vital-layer candidates identified following Stable Flow~\citep{avrahami2025stable}. The spatial responses vary substantially across diffusion depth despite using the same image and timestep.}
    \label{fig:layer_representation}
\end{figure}

We provide qualitative visualizations of the intermediate representations used by DiFuse. Figure~\ref{fig:timestep_representation} fixes the source layer and varies the denoising timestep, whereas Figure~\ref{fig:layer_representation} fixes the timestep and compares different diffusion layers. The corresponding quantitative analyses are reported in the main paper and Appendix~\ref{app:ablations} and are not repeated here.

\paragraph{Responses across timesteps.} Figure~\ref{fig:timestep_representation} visualizes the responses extracted from the fixed $g18$ layer at $t\in\{50,100,200,300\}$. Across the three examples, the representations retain spatially organized responses around foreground objects, object boundaries, and nearby regions, while their sharpness and spatial extent vary across timesteps.

\paragraph{Responses across diffusion layers.} Following the vital-layer analysis of Stable Flow~\citep{avrahami2025stable}, Figure~\ref{fig:layer_representation} visualizes the candidate layers at the fixed timestep $t=100$. Despite using the same image and timestep, different layers exhibit distinct spatial response patterns. These differences motivate the downstream source-layer screening in Appendix~\ref{app:source_layer_screening}, where the final source layer is selected according to multimodal reasoning performance.

These visualizations provide qualitative evidence that intermediate diffusion representations retain spatial organization, but should not be interpreted as attention maps or as direct explanations of individual predictions. The final timestep and source layer are determined by the quantitative results.

\section{Additional Qualitative Results}
\label{app:qualitative}

\begin{figure}[h]
    \centering
    \includegraphics[width=1.0\textwidth]{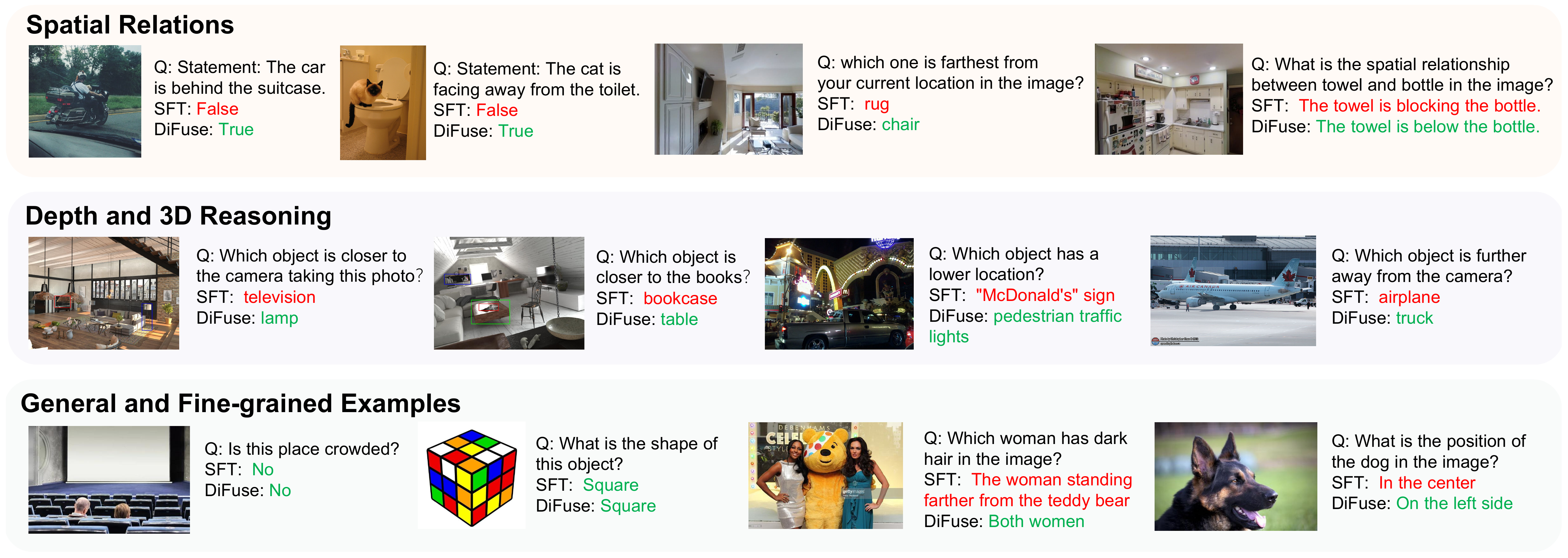}
    \caption{\textbf{Qualitative comparison between SFT and DiFuse.} Examples cover spatial relations, depth and 3D reasoning, and general and fine-grained visual understanding. Compared with the matched SFT baseline, DiFuse corrects several errors involving spatial relations and depth reasoning while largely preserving predictions on general visual understanding examples. Red and green responses denote incorrect and correct predictions, respectively.}
    \label{fig:qualitative}
\end{figure}

Figure~\ref{fig:qualitative} presents additional qualitative comparisons between the matched SFT baseline and DiFuse across different reasoning categories.

\paragraph{Qualitative comparison across reasoning categories.} The qualitative examples exhibit a selective pattern consistent with the quantitative results in Table~\ref{tab:main_controlled}. Compared with the matched SFT baseline, DiFuse primarily corrects errors on questions that require relational or geometric cues, including relative position, distance, and depth. In contrast, predictions that mainly depend on coarse scene recognition, object identity, or object shape are generally preserved. These observations are consistent with our hypothesis that native diffusion representations provide complementary structural information rather than uniformly altering the model's general multimodal behavior.

\end{document}

%% file: math_commands.tex
\usepackage{amsmath,amsfonts,bm}

\def\eqref#1{equation~\ref{#1}}

\def\1{\bm{1}}

\DeclareMathAlphabet{\mathsfit}{\encodingdefault}{\sfdefault}{m}{sl}
\SetMathAlphabet{\mathsfit}{bold}{\encodingdefault}{\sfdefault}{bx}{n}

